\def\year{2021}\relax
\documentclass[letterpaper]{article} 
\usepackage{aaai21}  
\usepackage{times}  
\usepackage{helvet} 
\usepackage{courier}  
\usepackage[hyphens]{url}  
\usepackage{graphicx} 
\usepackage{natbib}  
\usepackage{caption} 
\usepackage[english]{babel}
\usepackage{graphicx}
\usepackage{amsfonts}
\usepackage{amsmath}

\usepackage{mathrsfs}
\usepackage{xcolor}
\usepackage[normalem]{ulem}
\usepackage{esvect}
\usepackage{multirow}

\newtheorem{defn}{Definition}
\title{Not all users are the same: Providing personalized explanations for sequential decision making problems}
\author{
    Utkarsh Soni,
    Sarath Sreedharan,
    Subbarao Kambhampati
    \\
}
\affiliations{
School of Computing, Informatics, and Decision Systems Engineering,\\ Arizona State University, Tempe, AZ 85281 USA.
}

\begin{document}

\maketitle

\begin{abstract}
There is a growing interest in designing autonomous agents that can work alongside humans. Such agents will undoubtedly be expected to explain their behavior and decisions. While generating explanations is an actively researched topic, most works tend to focus on methods that generate explanations that are one size fits all. As in the specifics of the user-model are completely ignored. The handful of works that look at tailoring their explanation to the user's background rely on having specific models of the users (either analytic models or learned labeling models). The goal of this work is thus to propose an end-to-end adaptive explanation generation system that begins by learning the different types of users that the agent could interact with. Then during the interaction with the target user, it is tasked with identifying the type on the fly and adjust its explanations accordingly. The former is achieved by a data-driven clustering approach while for the latter, we compile our explanation generation problem into a POMDP. We demonstrate the usefulness of our system on two domains using state-of-the-art POMDP solvers. We also report the results of a user study that investigates the benefits of providing personalized explanations in a human-robot interaction setting.
\end{abstract}

\noindent Recent successes in AI have sparked great interest for autonomous agents to be deployed into our day to day life. Unfortunately, even the most powerful of AI systems fall quite short when it comes to the ability to work successfully with humans. A capability sorely missing in most of these systems is the ability to make sure that the agent acts in a manner that a human observer would find explainable. When failing to do so, it must have the ability to explain its decisions to their teammates.

While the problem of explaining AI decisions itself has been getting a lot of attention \cite{gunning2017explainable}, they have generally focused on generating explanations that are agnostic of the actual users of the system. While studies in social sciences have repeatedly shown the need to generate explanations that are social, i.e., explanations tailored to specific background of the explainee \cite{miller2019explanation}, most recent works in XAI have generally focused on generating either one-size-fits-all explanation or ones that are meant exclusively for the designers of the system. A recent trend in explanations, that have tried to look at the explainee's understanding of the task has been the works done under the umbrella of \textit{model reconciliation} \cite{chakraborti2017plan}. They generally look at the problem of explanation as one of correcting the human's mental model about the task so they can correctly evaluate the agent's decisions. Unfortunately these works either assume that the agent knows the exact model of the user (or some proxy of it)  (c.f. \cite{chakraborti2017plan, sreedharan2019model}) or consider cases which allow for possible uncertainty about the user model but require that the uncertain model be explicitly represented in a specific declarative forms (c.f \cite{sreedharan2018handling, sreedharan2018hierarchical}).

In the absence of explicitly given models, the agent would need to learn a new model for each new user. Realistically, this would be infeasible as learning a new model might require a lot of input from the user (not to mention the very act of interaction to learn these models could end up shifting their models). Moreover, the agent would not leverage the existing knowledge it has gained from learning the models of other users. We believe that a more reasonable setting is the one in which a given task will have categories of users such that the users in a category share a similar model of the task. We refer to these categories as user types. For example, for the task of robot-assisted cooking, one can imagine user types like novice, intermediate and expert chefs. Each type of user might require different explanations. Now if the robot has access to the models of each user type and knows the type of user it is interacting with, then it can successfully explain its behavior.

\begin{figure*} [!ht]
\centering
\includegraphics[scale=0.47]{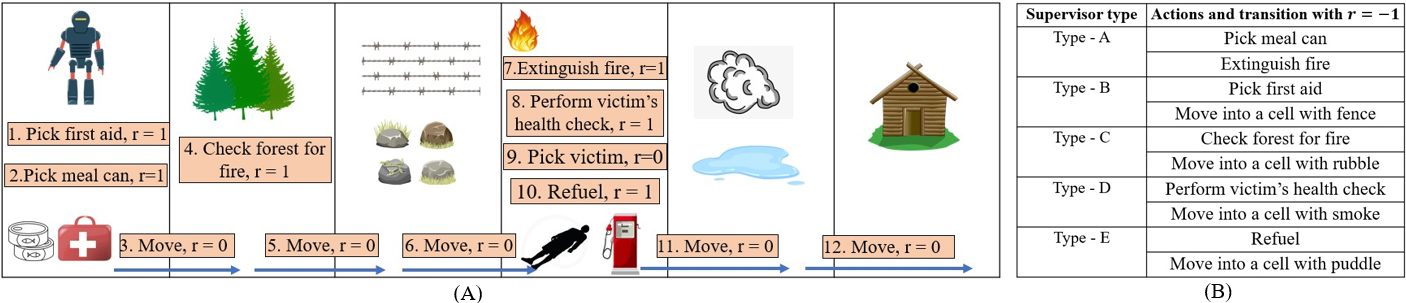}
\caption{(A) an instance of disaster-rescue domain along with the robot's execution trace (B) Inaccurate model information known to each user type}
\label{overall}
\end{figure*}

A significant challenge in such a setting comes from the fact that the user types might not be known in advance. In addition, even if they are, a user cannot be expected to know which type they belong to. This is particularly true in cases where each user may not be aware of all the classes of users or how these classes are typified. Given this complication, the system we design in this work is tasked with:
\begin{enumerate}
    \item Identifying the types of user that exists for the task and learning their mental model of the task.
    \item Inferring the type of user that the agent is interacting with, and using that to give personalized explanations. 
\end{enumerate}
In the following sections, we first introduce one of the planning domains that we use to evaluate our approach and then go over how an approximation of a user's model is learned. This is followed by a discussion on learning the model for each type of user. We then go over how our system handles explaining a plan trace to the user. Since the user's type is not known, we setup a POMDP based formulation that can reason about the hidden user's type and generate explanations accordingly. Finally, we test a variety of POMDP solvers on two planning domains and present a user-study that uses certain objective metrics to determine the benefits of providing personalized explanations.

\section{Illustrative example}
\label{illustrative_example}

We will use a disaster-rescue domain to illustrate the ideas of the paper. It is also one of the domains that we use to evaluate our solution approaches. The domain represents a scenario where a rescue robot is tasked with evacuating a victim to its shelter. There are also certain actions that the robot can do to get additional reward. Some example actions include picking up the first aid, extinguishing the fire, refueling at the fuel station, which give a positive reward. There are also certain obstacles like rubble and a puddle but the robot is capable of navigating through them. Thus, the robot doesn't incur a negative reward for navigating through any location with an obstacle. Figure \ref{overall} (A) shows one example instance of the domain and an optimal plan executed by the robot. Against each action, the corresponding reward is shown as well.

Outside the domain, we have a human supervisor whose task is to evaluate the robot's behavior. The supervisor can have  inaccurate knowledge of certain facts about the domain. As an example, we assume there are five types of supervisors identified as \textit{type-A} to \textit{type-E}. Each type of supervisor can have different knowledge about the domain. For instance, the supervisor of \textit{type-E} thinks that refueling is unnecessary and associates a negative reward with it. They also believe that the robot can break when moving into a location with puddle, and hence the corresponding transition has a negative reward in their model of the task. Figure \ref{overall} (B) enumerates the inaccurate facts each type holds for the domain. 

When evaluating the robot's plan trace, the supervisors of each type may find different parts of the plan inexplicable. For e.g., the supervisor of \textit{type-B} will be confused when the robot picks up the first-aid. Now, the robot isn't aware of all the supervisor types, let alone the type of the supervisor with whom it is interacting. One trivial way to make sure that the entire plan is explicable to the supervisor is to give them the entire model of the robot. This, however, can overwhelm them. We want the robot to provide only the required explanations to the supervisor. For this to be possible, the robot needs to figure out what are the types of supervisors that exists and what explanation messages would make the plan explicable for each type. Then, while interacting with the target supervisor, it needs to figure out their type and accordingly give personalized explanations.

\section{Background}

In this work, our focus is on robots that use discounted infinite horizon Markov Decision Processes (MDPs) for solving a sequential decision making problems \cite{russell2002artificial}. An MDP $\mathcal{M}$ can be formally defined as tuple $\langle S, A, T, R, \gamma \rangle$, where: $S$ is the set of states in the domain; $A$ is the set of actions that the robot can take in any state; $T$ is the transition function where $T(s, a, s')$ gives the probability that the robot will reach state $s'$ after taking action $a$ in state $s$; $R$ is the reward function such that $R(s, a, s')$ gives the reward for a transition $\langle s, a, s' \rangle$; $\gamma$ is the discounting factor for the rewards obtained (where $0 \leq \gamma < 1$). A policy $\pi$ for an MDP maps each state to a recommended action. The value for any state given $\pi$ is the expected cumulative discounted reward obtained if the robot follows $\pi$. For a given MDP $\mathcal{M}$, the optimal policy $\pi^{*}_{\mathcal{M}}$ is simply the policy that would give the highest value from any state. Lastly, the Q-value function, $Q_{\pi}(s, a)$, for an MDP gives the expected value obtained if the robot takes action $a$ in state $s$ and then follows the policy $\pi$. We use $Q^{*}$ to denote the Q-value function corresponding to the optimal policy for the MDP.

In many human-AI interaction scenarios, the model of the task used by the robot, $\mathcal{M}^r$ may differ from how the human understands the task (including the robot's capabilities). This discrepancy could imply that even when the robot follows optimal policies, the robot behavior may appear to be surprising and inexplicable to the human. Works like \cite{chakraborti2017plan} have looked at how to use explanations to address such scenarios. Specifically they consider revealing information about parts of the robot's model to correct the human's misunderstanding about the task. In this work, in particular, we use learned labeling models to figure out the specific model information to reveal to the user. The learned model functions as a proxy to the user's model of the task. Specifically, we assume a set of explanatory messages $\mathbb{E} = \{e_{1}, e_{2}, \dots, e_{n}\}$, that include information about the robot's model. An example message, for the disaster-rescue domain, would be \textit{the robot obtains a reward of 1.0 for picking up the first aid}.  Given a specific transition $\tau = \langle s, a, s'\rangle$, we learn a labeling model of the form $\mathcal{L}(\tau, \hat{\mathbb{E}})$, that maps the transition and a set of messages $\hat{\mathbb{E}}$ to either labels $1$ or $0$. Here $1$ refers to the fact that once the user is given a set of messages $\hat{\mathbb{E}}$ then they find the transition to be explicable, i.e., there exists an optimal policy in their updated model that can generate this transition. Note that our work accounts for the fact that different users of the system may have different backgrounds and mental model of the task. Hence, they may require different sets of explanatory messages. Our goal through this paper would be to see how we can set out to create an explanatory system that is able to create explanations that are tailor-made for specific users.

\section{Explanation in the Presence of Multiple User Types}

Our specific focus in this paper would be to handle cases where the robot needs to provide explanations to users with different backgrounds. To simplify the setting, we will assume that each user's understanding of the task could be mapped to one among a set of possible models, and thus maps to one of the possible labeling models in the set $\mathbb{L}$. Thus, for any user in the system of type $u_t$, we will denote the corresponding labeling function as $\mathcal{L}_{t} \in \mathbb{L}$. 

\subsection{Learning the set of labeling models}
\label{clustering}
Now the first step to setting up such an explanatory system for a task $\mathcal{M}^r$ and a set of messages $\mathbb{E}$ will be to learn the corresponding labeling function set $\mathbb{L}$. Thus, the system needs to identify all the user types, and learn their respective labeling functions. Similar to \cite{sreedharan2019model}, we will rely on data collected from users (referred to as observers) that include robot transitions along with messages and their corresponding labels provided by the observers. We will additionally assume that we know which observer id generated the label. Thus each datapoint takes the form  $d = \langle \tau, \hat{\mathbb{E}}, l, o\rangle$ where $\tau = \langle s,a,s'\rangle$ is the transition, $\hat{\mathbb{E}}$ is the message set, $l$ is the corresponding label and $o$ is the unique observer id. The learner's primary goal is to separate the collected data into $k$ groups where ideally, $k$ is equal to the true number of user types $N_{u}$(which is not known), and each group has data only from a single user type. This amounts to clustering the observers into $k$ such groups. 

To begin, each observer is represented as a vector $\dot{\mathbb{E}}_{o} \in [0, 1]^{|\mathbb{E}|}$ where each bit corresponds to some message in $\mathbb{E}$. For a given $\dot{\mathbb{E}}_{o}$, the set of messages whose bit is set to $1$ represents the smallest set of messages which when given to the observer $o$ will make every possible transition explicable to them. To obtain $\dot{\mathbb{E}}_{o}$, the learner first uses the datapoints $D_{o}$ collected from the observer to train their corresponding labeling function $\mathcal{L}_{o}$. Then, using $\mathcal{L}_{o}$, the smallest set of messages is computed that would make all the transitions in $D_{o}$ explicable to the observer. Finally, the obtained message set is used as an approximation of $\dot{\mathbb{E}}_{o}$. Note that in order to learn a $\mathcal{L}_{o}$ that is capable of computing the smallest message set, our data collection process is different than the one used in \cite{sreedharan2019model}. In our case, the data is collected from the observers via counterfactual queries of the form, if the message set $\mathbb{\hat{E}}$ was given to the user, then would they find a transition $\tau$ to be explicable. 

Now in this space of explanations, the observers belonging to the same type would tend to be closer to each other as they require similar explanations. Hence, the learner then applies clustering in this space given any input number of clusters $k$. Now that we have represented each observer in this space, the next step is to determine $N_{u}$. For this problem, we define an intra-cluster dissimilarity measure called \textit{disagreements} that evaluates a given clustering of observers. It is calculated as the total number of instances where two observers that have been clustered together and been given the same set of messages, disagree on the label of some transition. Formally we define \textit{disagreements} as follows:
\begin{defn}
For a given clustering of observers, \textit{disagreements} is equal to the total number of pairs of datapoints $d_{i} = \langle \tau_{i}, \hat{\mathbb{E}}_{i}, l_{i}, o_{i}\rangle$ and $d_{j} = \langle \tau_{j}, \hat{\mathbb{E}}_{j}, l_{j}, o_{j}\rangle$ that satisfy two conditions: first, $o_{i}$ and $o_{j}$ belong to the same cluster, and second, $\tau_{i} = \tau_{j}$, $\hat{\mathbb{E}}_{i} = \hat{\mathbb{E}}_{j}$, and $l_{i} \neq l_{j}$.
\end{defn}

We expect the value of \textit{disagreements} to be low when each cluster contains observers from a single type since they share a similar model of the task. Consequently, the learner applies clustering to the set of vectors $\dot{E}_{o}$ parameterized by the number of cluster $k$ where $k$ varies from $1$ to the total number of observers. For each clustering instance, the value of \textit{disagreements} is computed. At $k=1$, we expect the \textit{disagreements} to be highest. But as $k$ increase, observers of different types would ideally be placed in separate clusters thus decreasing \textit{disagreements}. When $k$ reaches $N_{u}$, each cluster should ideally have observers from a single type. Thus, further increasing $k$ would not reduce the \textit{disagreements} by too much (similar to the pattern followed by intra-cluster distance measures used in $k$-means clustering when increasing the number of clusters). We therefore speculate that the elbow point of the plot of \textit{disagreements} against increasing values of $k$ would be equal to the $N_{u}$. The learner programmatically finds this elbow point and uses it as the number of user types, $N_{u}$.

Given $N_{u}$ and its corresponding clustering output, the learner can now learn the labeling set $\mathbb{L}$. First, for each cluster, a union of all the datapoints for all the observer within that cluster is taken. Each of this set now contains datapoints corresponding to an identified user type $u_{t}$. The corresponding labeling model ${\mathcal{L}}_{t}$ is trained on the points in the set. This gives us the required set $\mathbb{L}$. Additionally, we learn a confidence measure which provides the confidence for the prediction given by any learned model. Specifically, for each ${\mathcal{L}}_{t}$ we learn a probabilistic model, $P_{{\mathcal{L}}_{t}}(l| \tau, \hat{\mathbb{E}})$ which gives the probability that the user of type $t$ will give $\tau$ the label $l$ provided they have been given the explanations $\hat{\mathbb{E}}$. For notational convenience, we define a set $U$ to include all the identified user types $u_{t}$.

\subsection{Explaining Behavior Traces}
\label{pomdp}
With the specific labeling models in place, we are now ready to generate explanations for a specific behavior trace $\Pi = \langle \tau_1,..., \tau_m\rangle$. An explanation in this setting consists of a sequence of explanatory messages of the form $\mathcal{E} = \langle \hat{\mathbb{E}}_1,..., \hat{\mathbb{E}}_m\rangle$ where the explanations $\mathbb{E}_{k}$ are given to the user at step $k$ of the interaction. At each step $k$, the user is presented with the prefix $\langle \tau_1, ..., \tau_k\rangle$ denoted as $\pi_{k}$, where the fact whether a user of type $u_t$ will find any transition $\tau_j$ to be explicable is given by the function $\mathcal{L}_{t}(\tau_j, \bigcup_{i=1}^k (\hat{\mathbb{E}}_i))$, i.e., the user would try to make sense of each transition using the current message as well as previous messages. The user is expected to provide labels for each transition in $\pi_{k}$.

Now each explanatory message may be associated with some communication cost. To allow for reasoning regarding the tradeoffs between communication and inexplicability, we will assume access to $C_{e}$, the function that gives the cost of presenting a set of messages $\hat{\mathbb{E}} \in \mathbb{E}$ to the user; $C_{inexp}$, the cost of showing an inexplicable plan step to the user, and $\lambda$, a constant value that decides the relative weight between the total communication and the total inexplicability cost for the entire interaction with the user. Thus the cost of any step in the explanation sequence becomes a linear combination of these two costs and we can define the total cost of a explanation sequence as follows:

\begin{defn}
The cost of an explanation sequence $\mathcal{E} = \langle \hat{\mathbb{E}}_1,..., \hat{\mathbb{E}}_m\rangle$, for a trace $\Pi$ and a user type $u_t \in U$ is given by 
\[C(\mathcal{E}) = \sum_{i=1}^{m}[\lambda.C_e(\hat{\mathbb{E}}_i) +  \sum_{j=1}^{i}(1 - \mathcal{L}_{t}(\tau_j, \bigcup_{k=1}^i (\hat{\mathbb{E}}_k)).C_{\text{inexp}}]\]
\end{defn}

Thus, if the user type were to be known, our goal would be to find a sequence that minimizes this total cost. Unfortunately, this is not true in our case and in addition to reasoning about the explanatory sequence, we would also need to reason about the hidden user types. We assume that all we have access to is a possible prior over the user types. A useful reasoning framework that we could leverage would be that of partially observable Markov decision process (POMDP) \cite{kaelbling1998planning}. We can cast this problem of reasoning about explanation into a meta POMDP of the form $\mathcal{M}_{\mathcal{E}} = \langle S_{\mathcal{E}}, A_{\mathcal{E}}, T_{\mathcal{E}}, O_{\mathcal{E}}, \Omega_{\mathcal{E}}, \gamma_{\mathcal{E}}\rangle$. Specifically, we turn the problem of identification of explanation with unknown user type into a POMDP where the user type is the hidden part of the state. Each of the components of $\mathcal{M}_{\mathcal{E}}$ are of the form:

\begin{itemize}
    \item $S_{\mathcal{E}}$:  Each state in the meta-POMDP would contain a prefix that is part of the trace to be explained, two separate sets of explanatory messages denoted as $\mathbb{E}_{\text{given}}$ and $\mathbb{E}_{\text{next}}$, a sequence $L$ of flags denoting whether the user found each of the transition within the prefix to be explicable or not given the messages in $\mathbb{E}_{\text{given}}$ and finally the hidden user type. The messages in $\mathbb{E}_{\text{next}}$ are the ones that will be presented to the user in the next interaction step. Note that the user type is unobservable to the robot. Formally, let $L_{\pi_{k}}$ denote the set of all possible sequence of explicability labels for the transitions within $\pi_{k}$. Then the state space is defined as:
    \begin{equation*}
    \begin{split}
        S_{\mathcal{E}} = & \{\langle \pi_{k}, L, \mathbb{\hat{E}}_{\text{given}}, \mathbb{\hat{E}}_{\text{next}}, t\rangle\ | k \in [1, m], L \in l_{\pi_{k}}, u_{t} \in U \\
        & \mathbb{\hat{E}}_{\text{given}} \in 2^{\mathbb{E}}, \mathbb{\hat{E}}_{\text{next}} \in 2^{\mathbb{E}} \text{ and } \mathbb{\hat{E}}_{\text{given}} \cap \mathbb{\hat{E}}_{\text{next}} = \emptyset \} \\
    \end{split}
    \end{equation*}
    \item $A_{\mathcal{E}}$: Actions in the POMDP correspond to the set of explanatory messages, $\mathbb{E}$, that the robot can present to the user in addition to a special action called \textit{explain}. Thus, $A_{\mathcal{E}} = \mathbb{E} \cup \{\text{\textit{explain}}\}$.
    
    \item $T_{\mathcal{E}}$: The transition function can be described as acting in two different modes. Say the interaction is in the state $s = \langle \pi_{k}, L, \mathbb{\hat{E}}_{\text{next}}, \mathbb{\hat{E}}_{\text{given}},u_{t} \rangle$. When the action, $a = \hat{e}$ (where $ \hat{e} \in \mathbb{E} - \mathbb{\hat{E}}_{\text{given}}$) then the next state is simply $\langle \pi_{k}, L, \mathbb{\hat{E}}_{\text{next}} \cup \hat{e}, \mathbb{\hat{E}}_{\text{given}}, u_{t} \rangle$ i.e. the explanatory message is added to the set of messages to be given in the next interaction step. In contrast, when $a = \text{\textit{explain}}$ or $a \in \mathbb{\hat{E}}_{\text{given}}$ (i.e. the robot chose an explanatory message that has already been given), then the transition function takes the interaction to its next step where the user is presented with the prefix $\pi_{k+1}$ and given the messages in $\mathbb{\hat{E}}_{\text{next}}$. In this case, there will be multiple possible next states where each state's label sequence correspond to a possible sequence of labels given to the $\pi_{k+1}$ by the user of type $u$. In all those states, the prefix is set to $\pi_{k+1}$, $\mathbb{\hat{E}}_{\text{given}}$ is set to $ \mathbb{\hat{E}}_{\text{given}} \cup \mathbb{\hat{E}}_{\text{next}}$ and $\mathbb{\hat{E}}_{\text{next}}$ is set to $\emptyset$ in that order. The user type is unchanged. Assuming each transition will be labelled independently of each other by any user, the transition probabilities for each state would be computed as $\Pi_{i=1}^{k+1}P_{\mathcal{L}_{t}}(l_{i}| \mathbb{\hat{E}}_{\text{given}})$ where $l_{i}$ is the label given to $\tau_{i}$.
    
    \item $R_{\mathcal{E}}$: The reward that the robot obtains will incorporate the cost associated with presenting an inexplicable transition, and the cost of communicating an explanatory message. For a transition $\langle s, a, s' \rangle$ with $\pi_{k}$ as the prefix in $s'$, the reward function is defined as:
    \begin{equation*}
        R_{\mathcal{E}} =
            \left\{
            	\begin{array}{ll}
            		-1.\lambda.C_{e}(a)  & \mbox{if } a \in \mathbb{E} \\
            		-1.\sum_{i=1}^{k} C_{inexp}.(1 - l_{i} )& \mbox{if } a = \text{\textit{explain}}
            	\end{array}
            \right.
    \end{equation*}
    \item $\Omega_{\mathcal{E}}$: The set of all observations would be all possible explicability labels for all prefixes $\pi_{k}$ in $\Pi$. In addition, $\Omega_{\mathcal{E}}$ will contain a special observation denoted as $l_{\phi}$.
    
    \item $O_{\mathcal{E}}$: The observation model is deterministic. If the interaction moves to its next step, then the observation is the label sequence $L$ associated with the next state. Otherwise, it is simply $l_{\phi}$. 
    
    \item $\gamma_{\mathcal{E}}$: We leverage discounting to allow for on-time explanations of the form studied in \cite{zakershahrak2019online}.
\end{itemize}

 Apart from the fact that we are explaining a sequence of traces, our embedding of the explanation reasoning into a sequential reasoning framework gives us additional advantages. For one, we are able to leverage the POMDP's ability to naturally generate information gathering strategies when appropriate to help proactively identify the user type. In general; for each observation $L$, it provides us information about the underlying user type. Assuming each transition is labelled independently, the belief about the user type is updated as follows in each interaction step: 
\begin{equation*}
\begin{split}
    b_{k+1}(t) = & (1/Z).P(L | \pi_{k+1}, \mathbb{\hat{E}}_{\text{given}}, t). b_{k}(t) \\
    = & (1/Z).\Pi_{i=1}^{k+1} P(l_{i} | \tau_{i}, \mathbb{\hat{E}}_{\text{given}}, t). b_{k}(t) \\
    = & (1/Z).\Pi_{i=1}^{k+1} P_{\mathcal{L}_{t}}(l_{i} | \tau_{i}, \mathbb{\hat{E}}_{\text{given}}).b_{k}(t) \\
\end{split}
\end{equation*}
Where $b_{k}(t)$ denotes the probability that the target user is of type $t$ at $k^{\text{th}}$ interaction step, $l_{i}$ the label provided by the user for the transition $\tau_{i}$ and $\mathbb{\hat{E}}_{\text{given}}$ the explanations that have been presented so far. To leverage the belief update, the POMDP policy could choose to employ explanation messages at a step that are designed to help reduce the possible uncertainty regarding the user's type. For example in a case with two user types, say a given transition is inherently explicable for one type of user while it requires an explanation for the other type. By withholding any explanation, the system would be able to tell the exact user type from the next observation. Another advantage of using the sequential reasoning framework is that it allows us to leverage the fact that explanations may be non-monotonic \cite{chakraborti2017plan}, i.e. the POMDP would prevent revealing unnecessary information up front that could potentially lead to the user getting confused.

While there exist exact algorithms for solving POMDPs, they are intractable for problems with long horizon (which in the case of $\mathcal{M}_{\mathcal{E}}$ is $|\Pi| + |\mathbb{E}|$). Instead we will mostly focus on using approximate POMDP solvers. In particular, we will consider approximations that are myopic, i.e. their ability to account for information gathering strategies is limited and we will also be looking at approximations that rely on at least some limited amount of lookahead. 

Although reasoning about information gathering is useful, it is also one of the main complexities in POMDPs \cite{fern2014decision}. Thus, using myopic solvers can be highly efficient. In general, we believe that myopic behavior will not adversely affect the cost of the interaction for our problem. Intuitively, this happens because unexpected labels can inform the robot about the actual user type. For instance, say the robot’s current belief is more probable to an incorrect user type and it gives an explanation specific to that user type for a transition. The actual user might still find that transition to be inexplicable, thus driving the belief to the correct type.  We will next go over the POMDP solvers that we use for generating explanations. 
\subsection{Myopic Approximation: QMDP}

QMDP \cite{littman1995learning} is a highly efficient way to learn a POMDP's policy that uses the Q-values of states in the underlying MDPs to approximate the Q-value of a belief state. If the user type of target user is known, then $\mathcal{M}_{\mathcal{E}}$, simply becomes an MDP. We use a variant of this MDP such that its optimal policy can be learned efficiently. The state in this MDP would consist of state components of $\mathcal{M}_{\mathcal{E}}$ not including the labels sequence and the user type. We ignore the labels to prevent the state space from blowing up in size (which would otherwise be exponential in length of the trace). To retain the robot's ability to prevent inexplicable labels, we instead, modify the reward function $R_{\mathcal{E}}$ for the case when $a = \text{\textit{explain}}$. In that case, for the MDP corresponding to a user type $t$, the reward value for transitioning into a state with prefix $\pi_{k}$ is the expected inexplicability cost of the form $-1.\sum_{i=1}^{k}C_{\text{\textit{inexp}}}.P_{\mathcal{L}_{t}}(l_{i} = 0 | \tau_{i}, \mathbb{\hat{E}}_{\text{\textit{given}}})$.  

Let $Q_{t}^{*}$ denote the Q-value function for the MDP corresponding to user type $t$ for the optimal policy. Then the Q-value for a belief state $b$ and an action $a$ can be approximated as 

\begin{equation}
\label{eq1}
Q(b, a) \approx \sum_{t} b(t).Q_{t}^{*}(s_{o}, a).
\end{equation}

where $s_{o}$ are the state components corresponding to the MDP for user type $t$. Intuitively, the Q-value is being computed for a scenario in which all the ambiguity of the user type would be resolved in the next step. Thus, the Q-value estimation doesn't account for the value provided by information gathering actions. 

\subsection{Myopic Approximation: QHR}
This is a variant of a method proposed in \cite{fern2014decision} to approximate the $Q_{t}^{*}$ values. Although ignoring the labels helps in reducing the state space, it still increases exponentially in the number of messages. Thus in the QHR approach, instead of computing the optimal $Q_{t}^{*}$, we will use the value of a suboptimal policy. The suboptimal policy assigns an action to any state assuming that the robot will not be allowed to provide explanations from the next interaction step. In this case, we let each action $a$ be a set of explanatory messages i.e. $a \in 2^{\mathbb{E}}$ and calculate the value of the suboptimal policy in closed form as follows
\begin{equation*}
\begin{split}
     Q_{t}^{approx}(s_{o}, a) = & -1*(\lambda.C_{e}(a) + \\ & \sum_{j=k}^{|\Pi|} \sum_{i=1}^{j} P_{\hat{\mathcal{L}}_{t}}(l_{i}=0 | \tau_{i}, \hat{\mathbb{E}}_{j} \cup a).C_{\text{inexp}}) \\
\end{split}
\end{equation*}
where $k$ is the current step in the interaction. The estimate can be computed much faster than computing the entire policy for the corresponding MDP. Once computed, the value can be used in to get the Q-value estimate $Q(b, a)$ for any belief state using equation \ref{eq1}. 

\subsection{POMCP with $d$-step lookahead}
The POMCP algorithm \cite{silver2010monte} has been shown to achieve high performance for POMDPs with large state space. We used POMCP on $\mathcal{M}_{\mathcal{E}}$ as the approximation that allows for look ahead. In our work, we made two modifications to the original POMCP approach. First, while the original algorithm uses particle filter algorithm to approximate belief update during the search, we performed exact belief update as described in an earlier section. Secondly, instead of always running the simulation until the terminal state, we only run the simulation till depth $d$ in the search tree. If the node corresponding to the history encountered till $d$ is not present in the tree, then the simulation proceeds with the rollout. Otherwise, the simulation is terminated and value for that belief node is computed using the QMDP technique.

\section{Evaluation}

\subsection{Computational experiments}

We validate our approach by applying it on the disaster-rescue domain and the four rooms domain \cite{sutton1999between}. The disaster-rescue domains had the $5$ types of users described in Figure \ref{overall}. For the four rooms domain, we vary the number of user types and their models were selected randomly. For evaluation, we define a model of the domain for each type as an MDP. The parameter values of the model will depend on the knowledge the user type has about the domain. For example, in the disaster-rescue domain the MDP for user type $B$ will have a negative reward for the transition in which the robot moves into a location with a fence. The MDP models are used to simulate users for any type for evaluation.

For the four rooms domains, we varied the total number of user types between $2$, $3$ and $4$. Thus, we had three different settings for the domain (as opposed to the single setting of $5$ user types for the other domain). For each setting in the four rooms domain, the model of each user type was randomly selected. First, we identify a parameter set $\mathcal{M}$ that defines the domain's MDP. It consists of the goal locations, discounting factor, step cost, special locations with some penalty, and magnitude of the penalty. We associate two different values with each parameter $i$ in $\mathcal{M}$ denoted as $\mathcal{M}^{r}_{i}$ and $\mathcal{M}^{t}_{i}$. We then instantiate the robot's model by setting its MDP parameters to their correponding values in $\mathcal{M}^{r}_{i}$. For each of the user type $t$, some model parameters $\mathcal{M}_{t} \subset \mathcal{M}$ are selected such that $\bigcup_{t} \mathcal{M}_{t} = \mathcal{M}$. Then each of the selected model parameters $i$ is assigned the value $\mathcal{M}^{t}_{i}$. The remaining MDP parameters are assigned the value $\mathcal{M}^{r}_{i}$.

For each setting within the domains, the data required to identify the user types that exist in the domain and learn their labeling models, is collected by having simulated users of each type label plan traces. We created $3$ observers for each type (collecting $100$ and $300$ points per observer for four rooms and disaster-rescue). For each observer, traces were generated where the robot starts at some random initial state and follows the optimal policy till it reaches a terminal state or a trace length of $10$. For each trace, a set of randomly selected explanatory messages from the set $\mathbb{E}$ are given to the user. The user's model is updated according to the messages and labels are obtained for each step of the trace. We label a transition to be explicable if the optimal policy in the updated user's model can generate that transition. 

Once the data is collected from the observers, we have access to a set of labeled transitions for each user type. This reflects the ideal clustering that can be achieved as data per type is perfectly separated. We learn the labeling models, $\mathbb{L}_{pc}$, corresponding to each user type. These models represent the best possible labeling models that can be learned for each user type. For a given set of labeled transitions, the labeling model is learned as a decision tree classifier. The input features to the decision tree consists of the transitions (where the state is defined by the grid locations of the robot for both the domains), and the set of explanatory messages. The output is the label given by the user. The test set accuracy after training the classifier is used as the confidence measure $P_{\mathcal{L}_{t}}$. The testing accuracies for the $\mathbb{L}_{pc}$ models across the domains for all types of users in each setting was $\geq 0.98$. 
 
To evaluate our setup for scenarios where the explanatory system has no prior knowledge about the user types, we apply the clustering strategy described earlier for the data collected for each setting. We used the approach presented in \cite{satopaa2011finding} to find the elbow point on the \textit{disagreements} vs $k$ plots which was taken as $N_{u}$. We then learn the labeling models for the output clusters which are assumed to be the labeling models for the identified user types in the domain. Our clustering approach was able to achieve ideal clustering i.e. it was able to find the correct number of user types, and was able to group observers of same type together in both domains for all the cases except for the setting with $4$ user types for four rooms domain. On analyzing the last case, we found that two user types were assigned to model parameters such that all the transitions given to their observers were explicable which is why the clustering strategy placed their observers in the same cluster. Finally, for the labeling models learned for the identified user types in each setting, we got testing accuracies similar to the $\mathbb{L}_{pc}$ models (i.e $\geq 0.98$).

For a baseline technique, for each setting, we learn a single labeling model for the entire set of labeled transitions collected from all the observers. This labeling model is then attributed to all the users in that setting. This amounts to assuming that there exists only a single user type. Since the labeling model of the target user is now known, the explanation can be generated easily by solving the problem as an MDP (as described before in QMDP algorithm).

To evaluate the baseline and the POMDP solvers, we define a regret measure, $\mathcal{R} = C(\mathcal{E}) - C(\mathcal{E})_{pc}$ where $C(\mathcal{E})$ is the cost of explanation sequence generated by the solver and $C(\mathcal{E})_{pc}$ is the cost of explanation sequence generated by an oracle agent which has been given the user's type and their labeling model in $\mathbb{L}_{pc}$. 

We compute the regret value under different communication and inexplicability trade-off conditions by varying $\lambda$. The cost of communicating an explanatory message and the cost of presenting an inexplicable transition to the user were both set to $1$ unit. For each value of $\lambda$, we generate $3$ traces where the robot start with a random initial state and follows the optimal policy till it reaches a terminal state. We then use our POMDP and baseline solvers to generate explanations for the traces for a user of each type for each setting across the two domains. The POMCP solver was used with a lookahead of $2$ steps. Table \ref{tab:1} shows the average regret measure  for both the domains for each value of $\lambda$. The average was taken across all the plan traces, and all the users within each setting in the domain. It also shows the average cost ($C(\mathcal{E})_{pc}$) for the oracle. The low regret values suggest that the our system is indeed able to figure out the user type and provide required explanations to the user. Its performance is quite close to the an oracle agent which has been given the best possible labeling model for the target user. Moreover, all of our techniques significantly outperform the baseline technique. Among our solvers, both QMDP and POMCP appear to have similar performance while outperforming QHR.

\begin{table}[!t]
  \centering
\scriptsize
  \begin{tabular}{|c|c|c|c|c|c|c|}
  \hline
  \multirow{2}{*}{Domain} & \multirow{2}{*}{$\lambda$} & \multirow{2}{*}{\shortstack[l]{Oracle's\\$C(\mathcal{E})_{pc}$}} &  \multicolumn{4}{|c|}{Regret ($\mathcal{R} = C(\mathcal{E}) - C(\mathcal{E})_{pc}$)}  \\ \cline{4-7}
  & & & QMDP & POMCP & QHR & Baseline \\
  \hline \hline
  \multirow{5}{*}{\shortstack[l]{Four\\rooms}} 
   & 0.5 & 0.27 & 0.46 & 0.59 & 0.62 & 7.11 \\
   & 1.0 & 0.55 & 0.62 & 0.70 & 0.96 & 7.40 \\
   & 1.5 & 0.83 & 0.66 & 0.61 & 1.29 & 7.62 \\
   & 2.0 & 1.11 & 0.70 & 0.74 & 1.62 & 7.85 \\
   & 2.5 & 1.31 & 0.74 & 0.75 & 2.03 & 8.14 \\
   \hline \hline
  \multirow{5}{*}{\shortstack[l]{Disaster\\rescue}} 
   & 0.5 & 0.86 & 1.06 & 2.0 & 3.46 & 11.93 \\
   & 1.0 & 1.73 & 0.93 & 0.93 & 5.86 & 12.06 \\
   & 1.5 & 2.6 & 1.0 & 1.0 & 6.53 & 12.2 \\
   & 2.0 & 3.46 & 1.2 & 1.06 & 5.13 & 12.06 \\
   & 2.5 & 4.33 & 1.4 & 1.25 & 5.36 & 11.73 \\
   \hline
  \end{tabular}
  \caption{Average regret and $C(\mathcal{E})_{pc}$ values for the domains}
  \label{tab:1}
\end{table}

\subsection{User study}

We conducted a user study to investigate the effects of providing personalized explanations on the human-robot interaction when the robot is trying to explain a trace to the human. We performed a between-subject study where the participants were provided explanations by one of two techniques: \textit{personalized} and \textit{conformant} for the disaster-rescue domain with the user types described in Figure \ref{overall}. 

\subsubsection{Explanation techniques} The personalized explanations were generated using the QMDP framework (with $\lambda$ set to $1$) where the participant was treated as the user whose type is unknown to the framework. As explained in our methodology, the actual labels provided by the participant were used to update the robot's belief about their type. For learning the labeling model set $\mathbb{L}$, in order to ensure maximal personalization, we assumed perfect clustering and collected around $1000$ data points each from $3$ simulated observers per type. The accuracies for learned models were $>0.99$. In the conformant explanation technique, the robot generates the explanations such that the presented prefix would be explicable to the user of any type. Conformant explanations were generated by using the MDP formalism described in the QMDP algorithm with the reward function now computing the expected inexplicabilty cost of the prefix computed across all the user types, $t$. Specifically, when $a = \text{\textit{explain}}$, the reward for the transition is given as $(-1.\sum_{t}\sum_{i=1}^{k}C_{\text{\textit{inexp}}}.P_{\mathcal{L}_{t}}(l_{i} | \tau_{i}, \mathbb{E}_{\text{\textit{given}}}))$.

\subsubsection{Tested hypothesis} We compare the interactions the participant have with the robot for the two explanation techniques along three dimensions by testing the following hypothesis:

\noindent \textbf{Hypothesis 1} Personalizing explanations lead to a shorter interaction time with the robot.

\noindent \textbf{Hypothesis 2} Personalizing explanations lead to higher explicability of the robot's behavior.

\noindent \textbf{Hypothesis 3} When given personalized explanation, users are able to retain the updated model of the robot at the end of the interaction.

\subsubsection{Procedure}
The user study was conducted completely over a web based interface. However, each participant was also on call with one of the authors in case they had queries about their task. The participant was only allowed to ask queries to the author before their interaction with the robot began. The participants didn't have to share their screen while performing the study, and were allowed to keep their video disabled. We asked the participant to keep their mic on at all time during the study. 

In the user study, each participant is assigned one of the user types of the disaster-rescue domain and an explanation technique. In the beginning of the study, the participant is provided with specific details about the domain that are known to the type assigned to the participant. They are then trained on their task in the study which is to evaluate the robot behavior. At each step of this evaluation, they need to label each transition they have seen so far as expected or unexpected based on both the information provided about the domain as well as any explanations that have been provided by the robot (we retain the labels from the previous step to reduce participant's efforts in case they don't want to change any label). The participants were told that explanations are meant to correct the description of the domain and that they must update their domain knowledge when given any explanation. We make sure the participant understand the task by having them label an action with and without explanation that was added to the domain for illustration. 

The training is followed by an eligibility test where we test whether the participant remembers the task description accurately. The participants are given a list of actions, and asked to choose whether the action has a reward $ \geq 0$ or not. Upon passing the test, the participant gets to move on to the evaluation part of the study. After the evaluation, we conduct a quiz to check whether they remember the updated task description. In the quiz, the participant has to choose the correct statements among a set of $4$ statements describing the domain. Two of those statements are inaccurate, while the other two reflect the domain parameters that would be updated by the explanations that the assigned user type would need in order to find the entire trace explicable. For each correct response, the participant got a score of $1$.

\subsubsection{Results}
\begin{table}[]
\centering
\scriptsize
\begin{tabular}{|l|l|l|l|}
\hline
Measure                            & Conformant & Personalized & p-value \\ \hline
Interaction time                   & 120.48     & 100.85       & 0.04    \\
Inexplicable transitions & 1.45       & 0.65         & 0.032   \\
Quiz score                         & 2.6        & 3.45         & 0.024   \\ \hline
\end{tabular}
\caption{Objective measures for the explanation techniques}
\label{tab:my-table}
\end{table}

We performed a between subject design with explanation types: \textit{personalized} and \textit{conformant} as the factors. We collected data from a total of $40$ participants ($10$ females, age ranging from $19$ to $31$, mean age$=23.9$) where each explanation type was assigned $20$ participants each. The $5$ user types of the domain were assigned uniformly among the $20$ participants for each explanation type. To test each hypothesis, we collected certain objective measures from the interaction. For the first hypothesis, we measured the total time the participant took for completing the robot's evaluation. For the second hypothesis, we used the number of transitions that remained inexplicable at the end of the interaction. For the final hypothesis, we used the score the participant got on the quiz. The table \ref{tab:my-table}) provides the average value of each measure across all participants for each explanation type (interaction time is in seconds). We performed two tailed t-test and found the difference between explanation types to be significant for all the measures (p-values reported in the Table \ref{tab:my-table}). As seen from the results, \textit{personalized} explanations generated from our method outperforms \textit{conformant} explanations on all the measures.

\section{Related work}

In this work our focus has been on generating explanations that try to resolve the user's confusion that arise from the user's incorrect or incomplete understanding of the task. Such explanations have generally been referred to as model reconciliation explanation in the literature \cite{chakraborti2017plan}. In fact, we build on a specific variant of model reconciliation explanation that relies on labeling models instead of exact models \cite{sreedharan2019model}. Model reconciliation forms only a type of explanations in the larger landscape of explanations \cite{xaip-landscape}. Another major form of explanatory information are the ones designed to provide inferential assistance to the users, for example works like \cite{seegebarth2012making, eiflernew,khan2009minimal} etc. There is no reason to believe that such explanatory messages could not be incorporated into the framework. Many of these messages are generated independent of specific user information and thus are completely compatible with our framework. Our methods should be able to easily identify when certain user groups would benefit from such messages.

With respect to user specific explanations, while there have been a number of works that have pointed out the need for explanations being tailored to the end user's background knowledge (c.f. \cite{miller2019explanation,zhou2020different}) there has been less number of works that have done this. Within the context of sequential decision-making the closest work we are aware of is \cite{sreedharan2018handling}. Though as mentioned earlier they still require the possible multiple models to be representable in the form of an incomplete planning model. Moreover, their user adaptive explanations require the ability for the robot to ask specific questions about user's knowledge which may not be available to the system. Similarly \cite{sreedharan2018hierarchical} assumes the user to belong to one of many types, but expect the user's model to be an exact state abstraction of the current task model. This requirement may not necessarily  be always met. Even outside explaining sequential decision making problems, we know of very few works that allow incorporation of user information. One possible contender is the work done under the umbrella of TCAV \cite{kim2017interpretability}, where the user can specify concepts that can be used to build the explanation. Even in this case, other than being able to build explanations in user's vocabulary it doesn't take into account any other information about the user's background.

Our own method to generate the explanations rely on a compilation POMDP. POMDP, even with the extremely high computational overhead has been a popular framework for many human aware application. This is in many ways due to the fact that incomplete knowledge is a part and parcel of many real-world human-AI interaction scenarios. An example was the use of POMDPs to create hand-washing assistants for people suffering from dementia \cite{hoey2010automated}. Our own use of various approximation is very much in keeping with techniques used by these earlier works (particularly those investigated in \cite{fern2014decision}). 

\section{Conclusion and future work}
This paper proposes a way in which a robot can generate personalized explanations for the human in the loop for the case where different users can have different mental models of the task. We argue that while one size fits all explanations are easy to generate they can easily overwhelm the user of the system with unnecessary communication. We develop an explanatory system that first identifies the user types that exist for a task, and then uses a POMDP based formulation that is capable of inferring the user's type (and hence their model) on the fly. We show its competence on a disaster-rescue domain and four rooms domain under different trade-off conditions between communication and inexplicability cost. We saw that in the cases considered, all our approaches are able to come quite close, in terms of interaction cost, to a comparable explanatory system that is operating with known user type. We also presented a user-study where we saw a clear advantage of personalizing explanations in terms of some objective metrics. We see this work as being just the first step, and going forward we would like to extend this framework to more general settings. For one, we would like to relax the assumption that all user belong to a fixed set of types. We aim to investigate if Bayesian non parametric techniques would be useful to address these limitations.

\bibliography{references}

\end{document}